\documentclass[letterpaper, 10 pt, conference]{ieeeconf}  

\IEEEoverridecommandlockouts                              
\title{\LARGE \bf
Audio-Visual Sentiment Analysis for Learning Emotional Arcs in Movies
}

\author{ \parbox{3 in}{\centering Eric Chu\\
        MIT Media Lab\\
        {\tt\small echu@mit.edu}}
        \hspace*{ 0.5 in}
        \parbox{3 in}{ \centering Deb Roy\\
       MIT Media Lab \\
        {\tt\small dkroy@media.mit.edu}}
}

\usepackage{url}  
\usepackage{subcaption}  
\usepackage{amsmath}	
\usepackage{changepage}		
\usepackage{mdframed}  
\usepackage{setspace} 
\usepackage{graphicx}

\usepackage{afterpage}
\newlength{\oldtextfloatsep}

\begin{document}

\maketitle
\thispagestyle{empty}
\pagestyle{empty}

\begin{abstract}

Stories can have tremendous power -- not only useful for entertainment, they can activate our interests and mobilize our actions. The degree to which a story resonates with its audience may be in part reflected in the emotional journey it takes the audience upon. In this paper, we use machine learning methods to construct emotional arcs in movies, calculate families of arcs, and demonstrate the ability for certain arcs to predict audience engagement. The system is applied to Hollywood films and high quality shorts found on the web. We begin by using deep convolutional neural networks for audio and visual sentiment analysis. These models are trained on both new and existing large-scale datasets, after which they can be used to compute separate audio and visual emotional arcs. We then crowdsource annotations for 30-second video clips extracted from highs and lows in the arcs in order to assess the micro-level precision of the system, with precision measured in terms of agreement in polarity between the system's predictions and annotators' ratings. These annotations are also used to combine the audio and visual predictions. Next, we look at macro-level characterizations of movies by investigating whether there exist `universal shapes' of emotional arcs. In particular, we develop a clustering approach to discover distinct classes of emotional arcs. Finally, we show on a sample corpus of short web videos that certain emotional arcs are statistically significant predictors of the number of comments a video receives. These results suggest that the emotional arcs learned by our approach successfully represent macroscopic aspects of a video story that drive audience engagement.  Such machine understanding could be used to predict audience reactions to video stories, ultimately improving our ability as storytellers to communicate with each other.\\

\end{abstract}

\section{Introduction} 
Theories of the origin and purpose of stories are numerous, ranging from story as ``social glue'', escapist pleasure, practice for social life, and cognitive play \cite{boyd2009origin, gottschall2012storytelling}.

However, not all stories produce the same emotional response. Can we understand these differences in part through the lens of \textit{emotional arcs}? Kurt Vonnegut, among others, once proposed the concept of  ``universal shapes'' of stories, defined by the ``Beginning--End'' and ``Ill Fortune--Great Fortune'' axes. He argued nearly all stories could fit a core arc such as the classic ``Cinderella'' (rise-fall-rise) pattern \cite{vonnegut_video}. 

It is well known that emotions play no small part in people's lives. We have seen emotional narratives as a convincing medium for explaining the world we inhabit, enforcing societal norms, and giving meaning to our existence \cite{massey2002brief}. It has even been shown that emotions are the most important factor in how we make meaningful decisions \cite{lerner2015emotion}.

There is also evidence that a story's emotional content can explain the degree of audience engagement. The authors of \cite{berger2012makes} and \cite{milkman2014science} examined whether the valence and emotionality was predictive of New York Times articles making the Times' most e-mailed list. Ultimately, they found that emotional and positive media were more likely to be shared.

Motivated by the surge of video as a means of communication \cite{index2011cisco}, the opportunities for machine modeling, and the lack of existing research in this area, we makes steps towards tackling these questions by viewing movies through emotional arcs. We deliver the following contributions:
\begin{itemize}
	\item \textbf{Datasets.} We introduce a Spotify dataset containing over 600,000 audio samples and features that can be used for audio classification. We also collect annotation data regarding the sentiment of approximately 1000 30-second movie clips. Both datasets will be made publicly available\footnote{https://sosuperic.github.io/a-darn-good-yarn-website}.
	\item \textbf{Modeling emotional arcs.} We train sentiment classifiers that can be used to compute audio and visual arcs. We also motivate a) the use of dynamic time warping with the Keogh lower bound to compare the shapes of arcs, and b) k-medoids as the algorithm for clustering.
	\item \textbf{Engagement analysis.} We provide an example in which a movie's arc can be a statistically significant predictor of the number of comments an online video receives.
\end{itemize}

\section{Related work}

The work is most closely paralleled by \cite{reagan2016emotional}, which analyzed books to state that ``the emotional arcs of stories are dominated by six basic shapes.'' Using text-based sentiment analysis, they use a singular value decomposition analysis to find a basis for arcs. The bases that explain the greatest amount of variance then form the basic shapes of stories.

In contrast to our computational approach, writers at Dramatica \cite{dramatica} have manually analyzed a number of books and films under the Dramatica theory of story, which has since been used to create software that can guide writers.

Research in sentiment analysis has primarily been text-based, with work ranging from short, sentence-length statements to long form articles \cite{thelwall2010sentiment, pang2008opinion}. There has been comparatively little work on images, with the Sentibank visual sentiment concept dataset \cite{borth2013large} being a prominent example.


Neural networks have been used in earlier research for tasks such as document recognition \cite{lecun1998gradient}. In recent years, deep neural networks have been successful in both the visual and audio domain, being used for image classification \cite{krizhevsky2012imagenet}, speech recognition \cite{hannun2014deep}, and many other tasks.



Outside of emotional arcs, there exists other research that applies computational methods to understanding story. For instsance, the M-VAD \cite{torabi2015using} and MovieQA \cite{tapaswi2016movieqa} datasets include a combination of subtitles, scripts, and Described Video Service narrations for movies, enabling research on visual question-answering of plot. 

\section{Overview} 

\begin{figure}[h!]
	\centering
	\includegraphics[width=0.8\linewidth]{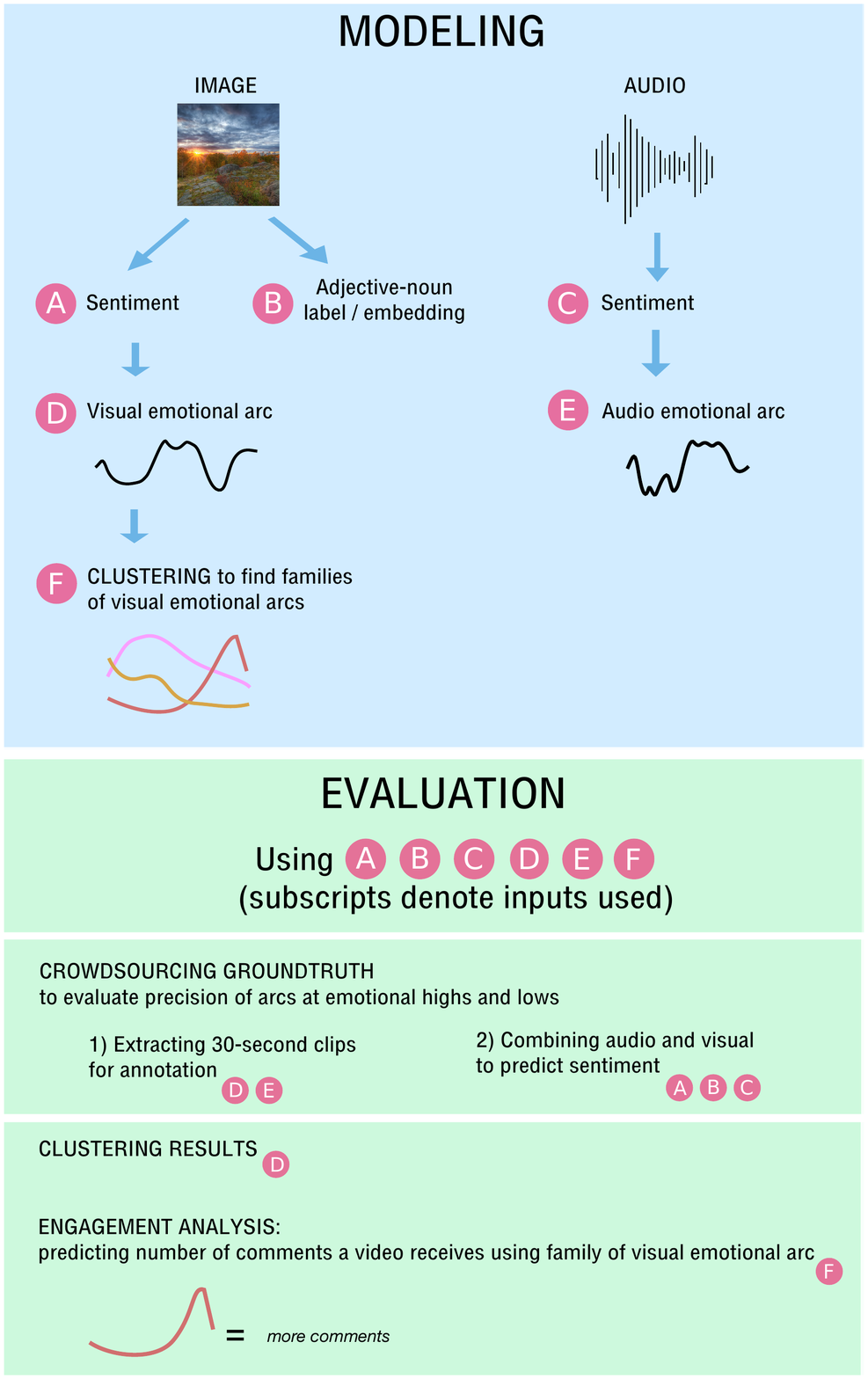}
	\caption{Overview}
	\label{overview_figure}
\end{figure}

Figure \ref{overview_figure} outlines the major pieces of this work. We note that modeling occurs at two scales -- micro-level sentiment, performed on a slice of video such a frame or snippet of audio, and macro-level emotional arcs.

Reflecting this distinction, we first evaluate the ability to accurately extract micro-level emotional highs and lows, which we refer to as \textit{emotionally charged moments} or emotional \textit{peaks and valleys}. Specifically, we measure precision as the amount of agreement in polarity between the models' sentiment predictions and annotators' ratings. Second, at the macro-level, we evaluate arcs by a) clustering, and b) using the arcs for the engagement analysis. We note that the final engagement analysis uses only the visual arcs, as we are currently limited by the amount of ground truth data that allows us to combine audio and visual predictions.

\section{Datasets}\label{sec_datasets}
\subsection{Videos -- Films and Shorts Corpora}
The system operates on two datasets collected for this work -- a corpora of Hollywood films and a corpora of hiqh quality short films. We selected films because they are created to tell a story. That there exist common film-making techniques to convey plot and elicit emotional responses also suggests the possibility of finding families of arcs.

Including Vimeo shorts allows us to 1) find differences in storytelling that may exist between films and newer, emergent formats, 2) possibly serve as a gateway to modeling, more generally, the short form videos that commonly spread on social networks, and 3) conduct an engagement analysis using the online comments left on Vimeo.


The first dataset, the \textit{Films Corpora}, consists of 509 Hollywood films. Notably, there is considerable overlap with the MovieQA \cite{tapaswi2016movieqa} and M-VAD \cite{torabi2015using} datasets. The \textit{Shorts Corpora} is a dataset of 1,326 shorts from the Vimeo channel `Short of the Week'. These shorts are collected by filmmakers and writers. A short can be 30 seconds to over 30 minutes long, with the median length being 8 minutes and 25 seconds.

\subsection{Image sentiment -- Sentibank dataset}
We use the Sentibank dataset \cite{borth2013large} of nearly half a million images, each labeled with one of 1,533 adjective-noun pairs. These pairs, such as ``charming house'' and ``ugly fish'', are termed \textit{emotional biconcepts}. Each biconcept is mapped to a sentiment value using the SentiWordnet lexicon.

\subsection{Audio sentiment -- Spotify dataset}
We started with the Million Song Dataset \cite{bertin2011million}, in which certain metadata, such as tempo or major/minor key, could in theory be used as a proxy for valence and other relevant target features. We also considered the Last.FM dataset of user-provided labels such as ``happy" or ``sad". Unfortunately, this dataset is significantly smaller.


To address our needs, we turned to the Spotify API. The API provides not only 30-second samples for songs, but also audio features used by the company. These include \textit{valence}, which measures the ``musical positiveness conveyed by a track.''
Other features include speechiness, liveness, etc.


\section{Methodology}
\subsection{Constructing emotional arcs}
Once the visual and audio models are trained for sentiment prediction (to be explained in Sections \ref{sec_image_model} and \ref{sec_audio_model}), each is applied separately across the length of the movie. To construct the visual arc, we extract a frame per second in the movie, resize and center crop it to size $256 \times 256$, and then pass it through the sentiment classifier. To construct the audio arc, we extract sliding 20-second windows.

\begin{figure}[h!]
	\centering   		\includegraphics[width=0.99\linewidth,height=0.25\linewidth]{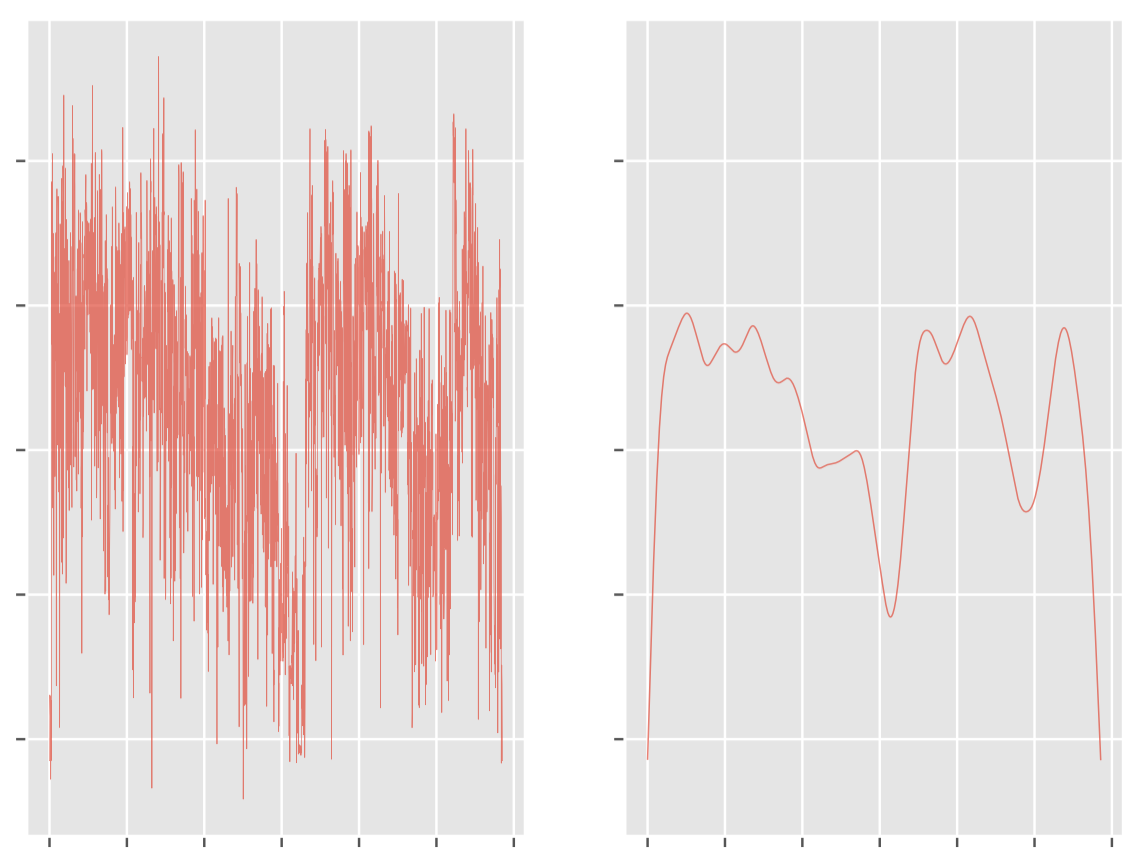}
	\caption{Effect of smoothing values for arcs: no smoothing versus $w = 0.1 * n$}
	
	\label{arc_smooth}
\end{figure}

\begin{figure}[h!]
	\centering
	\includegraphics[width=0.55\linewidth]{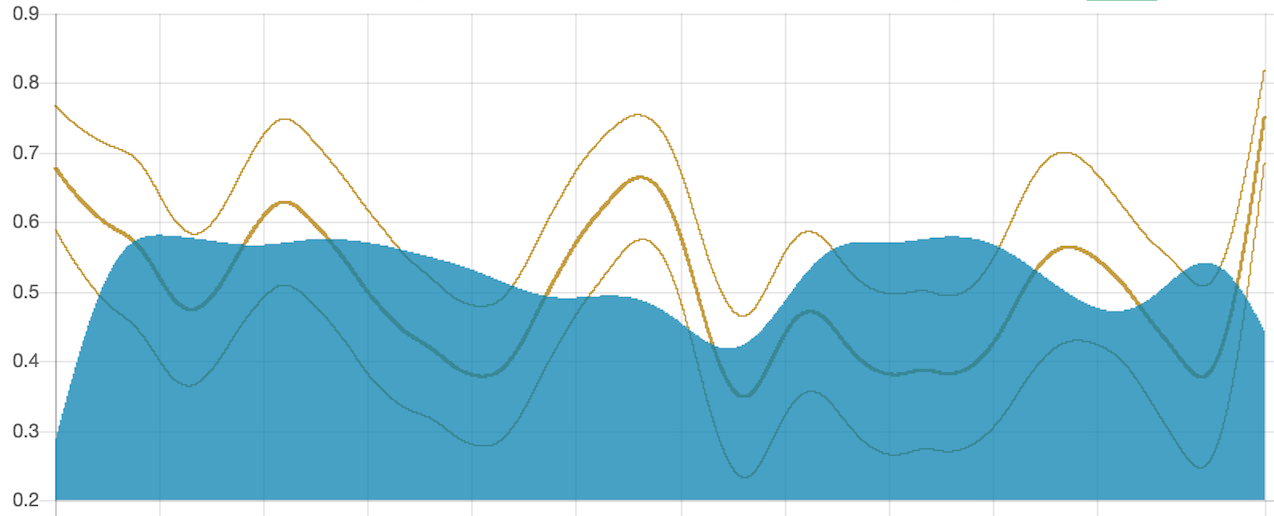}
	\caption{Audio (yellow) and visual (blue) arcs for \textit{Her}}
	\label{arc_audio_visual}
\end{figure}

While the left plot in Figure \ref{arc_smooth} shows that the macro-level shape is visible from the raw predictions, it is helpful to smooth these signals to produce clearer arcs by convolving each time series with a Hann window of size $w$. In downstream tasks, commonly used window sizes are 0.05, 0.1, and 0.2  * $n$, where $n$ is the length of the video. Example arcs are shown in Figure \ref{arc_audio_visual}, with the audio arc bounded by the confidence intervals to be described in Section \ref{sec_audio_model}.

\subsection{Image modeling}\label{sec_image_model}
The various effects of the visual medium has been well studied, ranging from the positive psychological effects of nature scenes \cite{ulrich1979visual} to the primacy of color, an effect so powerful that some filmmakers explicitly map color to target emotions in pre-production colorscripts \cite{amidi2015art}. We thus built models take a frame as input.

\subsubsection{Model}
We use a deep convolutional neural network based on the AlexNet architecture \cite{krizhevsky2012imagenet} to classify images. While a more state-of-the-art architecture would have higher accuracy, our focus was on building higher order arcs, for which this relatively simple model sufficed. However, we did use the PReLU activation unit, batch normalization, and ADAM for optimization to reflect recent advancements.

\subsubsection{Sentiment prediction}

The network was trained using images with a sentiment greater than 0.5 as `positive', and those less than -0.5 as `negative'. We used a learning rate of 0.01, a batch size of 128, and a batch normalization decay of 0.9. The performance is shown in Table \ref{sent_perf}.

\begin{table}[h!]
	\centering
    \scalebox{0.7}{
	\begin{tabular}{ |c|c|c|c| }
		\hline
		Accuracy  & Precision  & Recall & F1 \\\hline 
		0.652   &   0.753 &  0.729  & 0.741 \\\hline
	\end{tabular}
    }
	\caption{Performance of sentiment classifier}
	\label{sent_perf}
\end{table}

\subsubsection{Emotional biconcept prediction}
Using only the sentiment value is useful for creating emotional arcs, but it also discards information. Thus, we trained a second network that treats the biconcepts as labels. This network proves useful in creating \textit{movie embeddings} that broadly capture a movie's emotional content. Details are discussed in section \ref{sec_cf_combined_model}.

Only biconcepts with at least 125 images were used, leaving 880 biconcepts. The accuracy is shown in Table \ref{bc_perf}. Top-$k$ accuracy is defined as the percent of images for which the true label was found in the top-$k$ predicted labels. We also show the top-$k$ accuracy defined by whether the true adj / noun was found in the top-$k$ predicted labels.

\begin{table}[!htb]
	\centering
	\begin{minipage}{.49\linewidth}
		\centering
    	\scalebox{0.7}{
		\begin{tabular}{ |c|c| }
			\hline
            & Acc. \\\hline 
			Top-1 & 7.4\% \\\hline 
			Top-5 & 19.9\% \\\hline 
			Top-10 & 28.4\% \\\hline
		\end{tabular}
        }
		\subcaption{Predicting adj-noun pair}
	\end{minipage}
	\begin{minipage}{.49\linewidth}
		\centering
    	\scalebox{0.7}{
       	\begin{tabular}{ |c|c|c| }
			\hline
            & Match adj & Match noun \\\hline 
			Top-1 & 12.0\% & 15.1\% \\\hline 
			Top-5 & 30.2\% & 31.7\% \\\hline 
			Top-10 & 41.5\% & 40.7\% \\\hline
		\end{tabular}
        }
		\subcaption{Matching adj / noun in predicted adj-noun pair}
	\end{minipage}
	\caption{Performance of emotional biconcept classifier}
	\label{bc_perf}
\end{table}



\subsection{Audio modeling}\label{sec_audio_model}
Imagine `watching' a movie with your eyes closed -- you would likely still be able to pinpoint moments of suspense or sadness. While often secondary to the more obvious visual stimuli, sound and music can be played with or in contrast to the visual scene. With the idea that just a few seconds is enough to set the mood, we created a model for sentiment classification that operates on 20-second snippets of audio.

\subsubsection{Model}
We represent each audio sample as a 96-bin mel-spectrogram. We adopt the architecture used for music tagging in \cite{choi2016automatic}, which uses five conv layers with ELU and batch normalization, followed by a fully connected layer. 


\subsubsection{Sentiment prediction}
We used all samples that have a valence either greater than 0.75 or less than 0.25, leaving $\sim$200,000 samples. The performance is shown in Table \ref{audio_sent_perf}.

\begin{table}[h!]
	\centering
    \scalebox{0.7}{
	\begin{tabular}{ | c | c | c | c | } \hline 
		Accuracy & Recall & Precision & F1 \\\hline
		0.896 & 0.871 & 0.931 & 0.900 \\\hline
	\end{tabular}
    }
	\caption{Performance of audio sentiment classifier}
	\label{audio_sent_perf} 
\end{table}

\subsubsection{Uncertainty estimates}
Unfortunately, we face the problem of covariate shift, where the audio found in movies will often contain sound not found in the song-based training set. For example, there may be significant sections of background noise, conversation, or silence.

To handle `unfamiliar' inputs, we aim to produce confidence intervals for every prediction. While the softmaxed activations can be interpreted as probabilities, and hence a reflection of confidence, these probabilities can often be biased and require calibration \cite{niculescu2005predicting}. We thus follow a method introduced in \cite{gal2016dropout}, which produces approximate uncertainty estimates for any dropout network by passing the input $m$ times through the network with dropout \textit{at test time}, and using the standard deviation of the predictions to define a confidence interval around the mean of the predictions.

\subsection{Finding families of emotional arcs}\label{sec_cluster}

\subsubsection{Approach: k-medoids and dynamic time warping}
\begin{figure}[h!]
	\centering
	\includegraphics[width=0.5\linewidth]{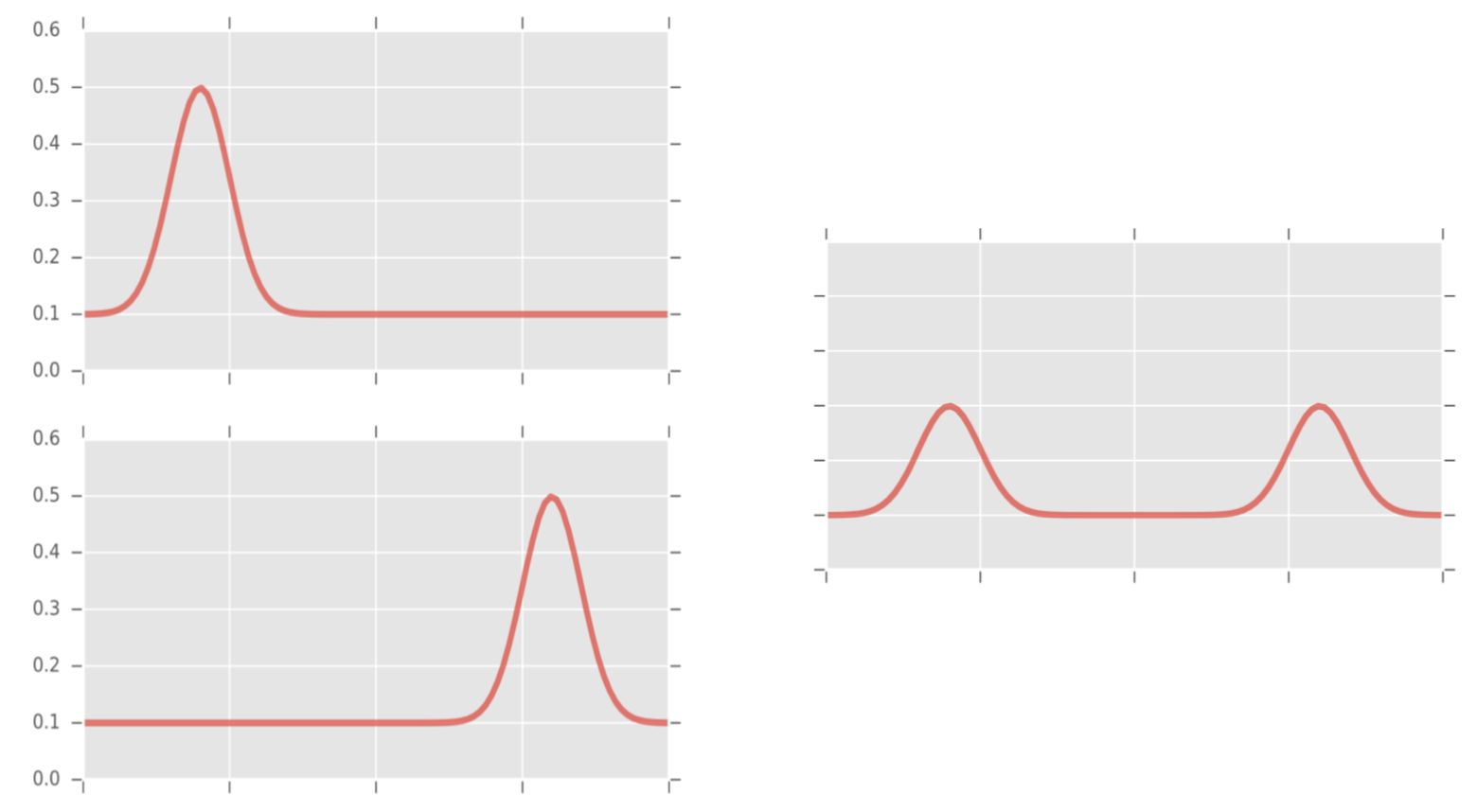}
	\caption{Pathological example of how k-means and Euclidean distance fails for clustering emotional arcs}
	\label{dtw_pathological}
\end{figure}

A naive approach to clustering arcs could be to use a popular algorithm such as k-means \cite{macqueen1967some} with an Euclidean metric to measure the distance between two arcs. However, this is a poor approach for our problem for two reasons:

\begin{itemize}
	\item Taking the \textit{mean} of arcs can fail to find centroids that accurately represent the shapes in that cluster. Figure \ref{dtw_pathological} shows a pathological example of when this occurs. The mean of the left two arcs has two peaks instead of one.
	\item The Euclidean distance between two arcs doesn't necessarily reflect the similarity of their shapes. While the left two arcs in Figure \ref{dtw_pathological} are similar in shape (one large peak), their Euclidean distance may be quite large.
\end{itemize}

With these limitations in mind, we turn to k-medoids \cite{kaufman1987clustering} with dynamic time warping (DTW) \cite{ding2008querying} as the distance function. K-medoids updates a medoid as the point that is the \textit{median} distance to all other points in the cluster, while DTW is an effective measures the distance between two time series that may operate at different time scales.

Given two time series $A$ and $B$ of length $n$, we construct a $n$x$n$ matrix $M$, where $M[i][j]$ contains the squared difference between $A_i$ and $B_j$. The DTW distance is the shortest path through this matrix. In Figure \ref{dtw_pathological}, for example, the DTW distance between the left two time series is 0.


\subsubsection{LB-Keogh for speed-up and better modeling}

Several techniques to speed up DTW center around creating a `warping window' that limits the available paths through the matrix $M$. The Keogh lower bound creates upper and lower bounds that envelop the original time series $A$, defined as:

\vspace{-0.5em}
\footnotesize
\begin{align}
	U_i = max(A_{i-r} : A_{i+r}) \\
	L_i = min(A_{i-r} : A_{i+r})
\end{align}
\normalsize
where $r$ is a parameter \textit{reach} that controls the size of the window. Intuitively, this controls how much a time series is allowed to warp. The lower-bounded $LB_{Keogh}$ distance between $A$ and $B$ is then given as:

\footnotesize
\begin{align}
	LB_{Keogh}(A,B) = \sqrt{\sum_{i=1}^n
		\begin{cases}
		(B_i - U_i)^2 & \text{if $B_i > U_i$} \\
		(B_i - L_i)^2 & \text{if $B_i < L_i$} \\
		0 & \text{otherwise}
		\end{cases}
	}
\end{align}
\normalsize
If $B$ lies inside the envelope of $A$, then the distance is 0.

Importantly, this approach has ramifications beyond increased speed up. Consider again the left two arcs in Figure \ref{dtw_pathological}. While both characterized by a large peak, it's possible that the second, \textit{ending} on a emotional moment, has greater impact. Consequently, we would like to only allow warping to a certain extent. The window does exactly that.

\subsubsection{Practical notes}
First, we exclude movies longer than $m$ seconds (10000 and 1800 for the Films and Shorts Corpora). After all, a 60 minute movie is hardly a `short'. Second, since we are interested in the overall \textit{shape} of the arc, we z-normalize each emotional arc. Finally, we settled on $r = 0.025 * n$.  This is close to $0.03 * n$, found to be optimal for a number of different tasks \cite{ratanamahatana2004everything}. 

\section{Evaluation -- Micro-level Moments}\label{sec_cf}

We evaluated the system's micro-level accuracy by its \textit{precision} in extracting emotionally charged moments from the emotional arcs. Notably, collecting ground truth data also allows us combine the audio and visual predictions.



Annotating an entire film, let alone multiple films, would be too costly and time intensive. We thus extracted video clips at peaks and valleys in the audio and visual arcs, hoping that these clips would lead to more interesting and informative annotations. Workers were asked to watch a clip and answer four questions regarding its emotional content. Each clip was annotated by three workers. 
\subsection{Crowdsourcing experiment}
We chose the CrowdFlower platform for its simplicity and ease of use. Answers to Question 1 (How positive or negative is this video clip? 1 being most negative, 7 being most positive) are referred to as \textit{valence ratings}. We define a \textit{positive rating} as a valence rating greater than 4, and a \textit{positive clip} as those with a mean rating greater than 4. Negative ratings and clips are similarly defined.
\subsection{Precision of system}
\subsubsection{Defining precision}
We define the precision as:

\footnotesize
\begin{equation}
\frac{\text{$|$peak \& positive$|$ + $|$valley \& negative$|$}}{\text{number of clips}}
\end{equation}
\normalsize

In other words, a clip was accurately extracted if a) it was extracted from a peak in either arc, and it was labeled as a positive clip, or b) it was extracted from a valley in either arc, and it was labeled as a negative clip.

\subsubsection{Overall precision}
Table \ref{cf_prec_all} lists the precision on both the full dataset and the full dataset with ambiguous clips (receiving both a positive and a negative rating) removed. We note that random chance would be 3/7 = 0.429. We argue that ambiguous clips should not be included, as it is unclear what their valence might be without more annotations. Further numbers are calculated with ambiguous clips removed.

\begin{table}[h!]
	\centering
    \scalebox{0.7}{
	\begin{tabular}{ |c|c| }
		\hline
		Set & Precision \\\hline 
		All clips &  0.642 \\\hline
		No ambiguous clips & \textbf{0.681} \\\hline
	\end{tabular}
    }
	\caption{Precision of clips: overall}
	\label{cf_prec_all}
\end{table}

\subsubsection{Precision of audio}
Next, we look at the precision of clips extracted from the audio arc, examining our hypothesis that predictions with smaller confidence intervals would be more accurate. This would help affirm our uncertainty estimates approach. Table \ref{cf_audio_prec} shows that smaller confidence intervals do indeed correspond with greater precision.
\begin{table}[h!]
	\centering
    \scalebox{0.7}{
	\begin{tabular}{ |c|c|c| }
		\hline
		Stddev & Audio-peak precision & Audio-valley precision \\\hline 
		[0, 0.02) & 4 / 4 = \textbf{1.0} & 81 / 88 = \textbf{0.921} \\\hline
		[0.02, 0.04) & 11 / 11 = \textbf{1.0} & 38 / 56 = \textbf{0.679} \\\hline
		[0.04, 0.06) & 28 / 40 = \textbf{0.7} & 21 / 35 = \textbf{0.6} \\\hline
		[0.06, 0.08) & 39 / 60 = \textbf{0.65} & 13 / 21 = \textbf{0.619} \\\hline
		[0.08, 0.1) & 43 / 68 = \textbf{0.632} & 8 /23 = \textbf{0.615} \\\hline
	\end{tabular}
    }
	\caption{Precision of clips extracted from audio emotional arc: smaller confidence intervals are more precise}
	\label{cf_audio_prec}
\end{table}

\subsubsection{Precision of various cuts}\label{sec_prec_cuts}
The precision on various subsets is shown in Table \ref{cf_prec_cuts_genre} (a). We highlight that the visual-peaks have low precision. This is explored in the next section and used to motivate feature engineering for the combined model in Section \ref{sec_cf_combined_model}.

\begin{table}[!htb]
	\begin{minipage}{.4\linewidth}
		\centering
    	\scalebox{0.7}{
		\begin{tabular}{ |c|c| }\hline
			Cut & Precision \\\hline 
			Audio-peaks  &  0.683 \\\hline
			Audio-valleys  & 0.758 \\\hline
			Visual-peaks  &  \textbf{0.508} \\\hline
			Visual-valleys  & 0.757 \\\hline
		\end{tabular}
        }
		\subcaption{Cuts}
	\end{minipage}%
	\begin{minipage}{.55\linewidth}
		\centering
    	\scalebox{0.7}{
		\begin{tabular}{ |c|c|c| }
		\hline
		Genre & Overall & Visual-peak \\\hline
		Action & 0.678 & \textbf{0.264} \\\hline
		Science Fiction & 0.699 & \textbf{0.333} \\\hline
		Thriller & 0.678 & \textbf{0.382} \\\hline
		Adventure & 0.726 & \textbf{0.443}  \\\hline
		Drama & 0.660 & \textbf{0.520} \\\hline
		Fantasy & 0.769 & \textbf{0.590} \\\hline
		Comedy & 0.705 & \textbf{0.667} \\\hline
		Animation & 0.798 & \textbf{0.667} \\\hline
		Family Film & 0.760 & \textbf{0.722} \\\hline
		Romance & 0.678 & \textbf{0.757} \\\hline
		Romantic Comedy & 0.677 & \textbf{0.823} \\\hline
	\end{tabular}
        }
		\subcaption{Genres}
	\end{minipage}
	\caption{Precision of clips: cuts and genre}
	\label{cf_prec_cuts_genre}
\end{table}




\subsubsection{Precision by genre}\label{sec_genre_prec}
Finally, we use \cite{bamman2014learning} to tag each movie with genres. A subset of the results is shown in Table \ref{cf_prec_cuts_genre} (b). The relatively poor precision of visual-peaks, as noted in the previous section, appears to be a product of poor precision on a number of genres. We also find a natural grouping of the genres when listed in this order. Genres with high visual-peak precision appear to be lighter films falling in the romance and family film genres.


Manual inspection of `incorrect' visual-peak clips from the action-thriller genres shows many scenes with gore and death, images unlikely to be found in the Sentibank dataset, which was culled from publicly available images on Flickr.


%
%

\subsection{Combined audio-visual model}\label{sec_cf_combined_model}
We create a linear regression model to predict the mean valence rating assigned by the annotators. In addition to standard features related to the clip's audio / visual valence (relative to the movie's mean, relative to the movie's max, etc.), we create two key features detailed below.



\textbf{Peakiness.} The function $p(a, i, r)$ approximates the slope and mean around a given point $i$ for arc $a$ (audio or visual), where $r$ is the window size around $a_i$. $p$ returns 4 values: $a_{i-1} - a_{i-r}$ (proportional to the slope left of the $i$), $a_{i+r} - a_{i+1}$ , $mean(a_{i-r:i-1})$ (the mean value left of $i$), and $mean(a_{i+1:i+r})$. This covers peaks, valleys, and inflection points. We use $r=0.025$ for our analyses.

\textbf{Movie embedding.} Motivated by the impact of genre in Section \ref{sec_genre_prec}, we sought to loosely summarize a movie's emotional content. We represent each frame by the penultimate activation of the biconcept classifier described in Section \ref{sec_image_model}. Next, we average these activations across 10\% chunks of the movie, resulting in a $10 \times 2048$ matrix. To translate these \textit{movie embeddings} to features, we take the mean of each 2048-sized vector, ending with a final $10 \times 1$ feature vector. While not shown in the interest of space, clustering these movie embeddings shows correspondence between clusters and genres, with romance, adventure, fantasy, and animated films being clearly visible.




\subsection{Precision of combined audio-visual model}

The performance of the final combined model is shown in Table \ref{cf_combined_prec}, along with various ablations of important features. Using all features, we achieve a precision of 0.894.

\begin{table}[h!]
    \scalebox{0.7}{
      \begin{tabular}{ |c|c|c|c|c|c| }
          \hline
          Feature set & Overall & Aud-peak & Aud-valley& Vis-peak & Vis-valley \\\hline 
          \textbf{All features} & \textbf{0.894} & 0.940  & 0.884 & 0.872 & 0.886  \\\hline
                    No \textit{peakiness} & 0.815 & 0.836 & 0.828  & 0.765  & 0.824 \\\hline
          No \textit{movie-embedding} & 0.784 & 0.869 & 0.786 & 0.722 & 0.752 \\\hline

		\end{tabular}
    }
	\caption{Precision of combined audio-visual model}
	\label{cf_combined_prec}
\end{table}



\section{Evaluation -- Macro-level}
Results shown here are based on the \textit{visual} arcs, which we consider to be the primary medium. We could in theory use arcs constructed from the \textit{combined} audio-visual model described in Section \ref{sec_cf_combined_model}. Experiments clustering the combined arcs, however, produce largely indistinct clusters. This is a result of a) the sparsity of ground truth data covering the entire audio-visual space found in movies (e.g. no clips were extracted from moments that were neutral in both the audio and visual arcs), and b) simply the small size of the dataset, leading to inaccurate combining of audio and visual.

\subsection{Cluster results}
\begin{figure}[h!]
	\centering
	\begin{subfigure}[t]{0.4\linewidth}
		\centering
		\includegraphics[width=\linewidth]{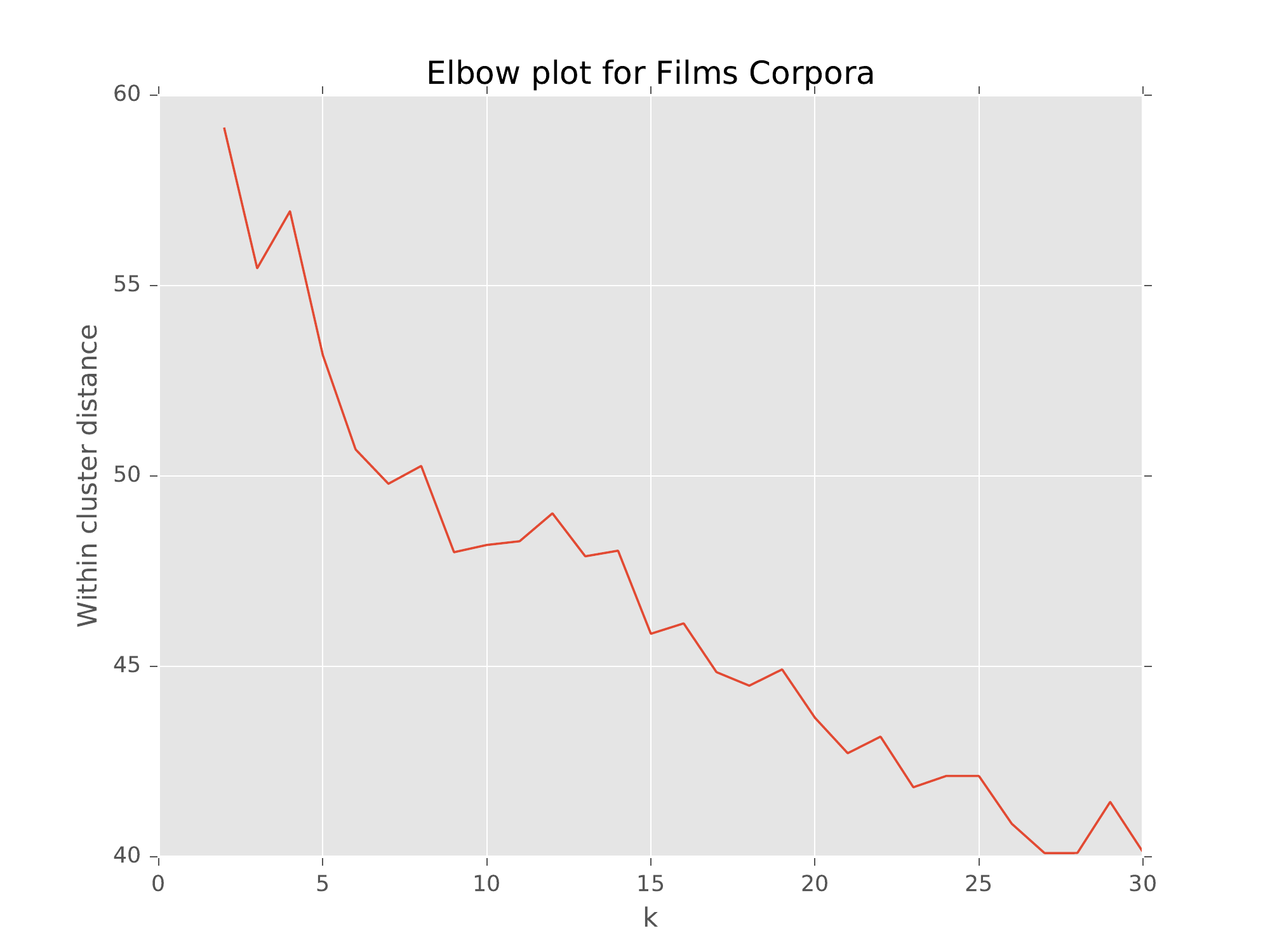}
		\caption{Films Corpora}
	\end{subfigure}
	\begin{subfigure}[t]{0.4\linewidth}
		\centering
		\includegraphics[width=\linewidth]{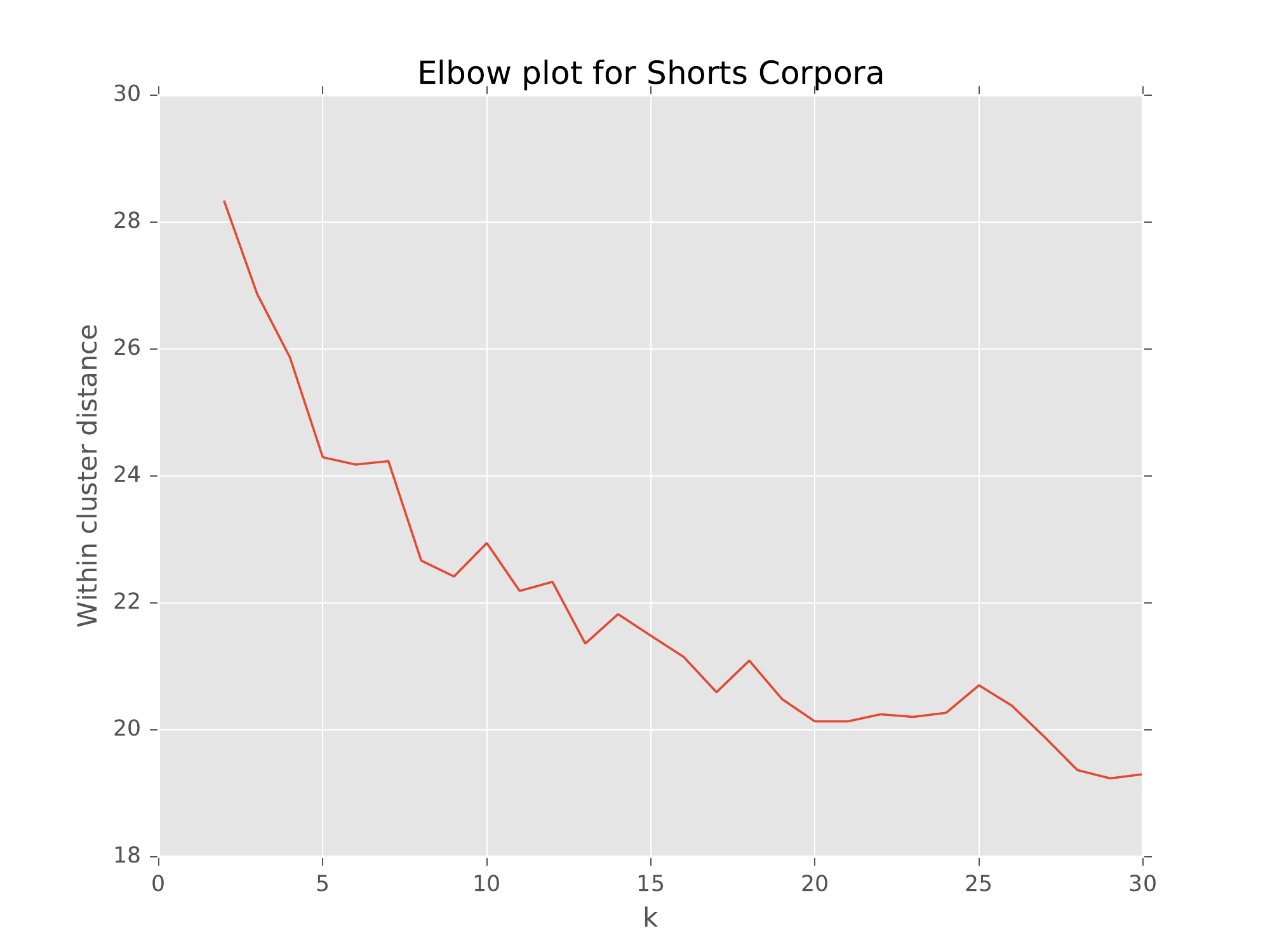}
		\caption{Shorts Corpora}
	\end{subfigure}
	\caption{Elbow plots for k-medoid clustering}
	\label{kmedoids_error}
\end{figure}

Figure \ref{kmedoids_error} shows the results of the elbow method, which plots the number of clusters $k$ against the within cluster distance (WCD). We can see possible `elbows' at $k$=5 and $k$=9. We briefly note that a k-means approach produced indistinct clusters and no discernible decrease in the WCD.

\begin{figure}[h!]
	\centering
	\includegraphics[width=0.55\linewidth]{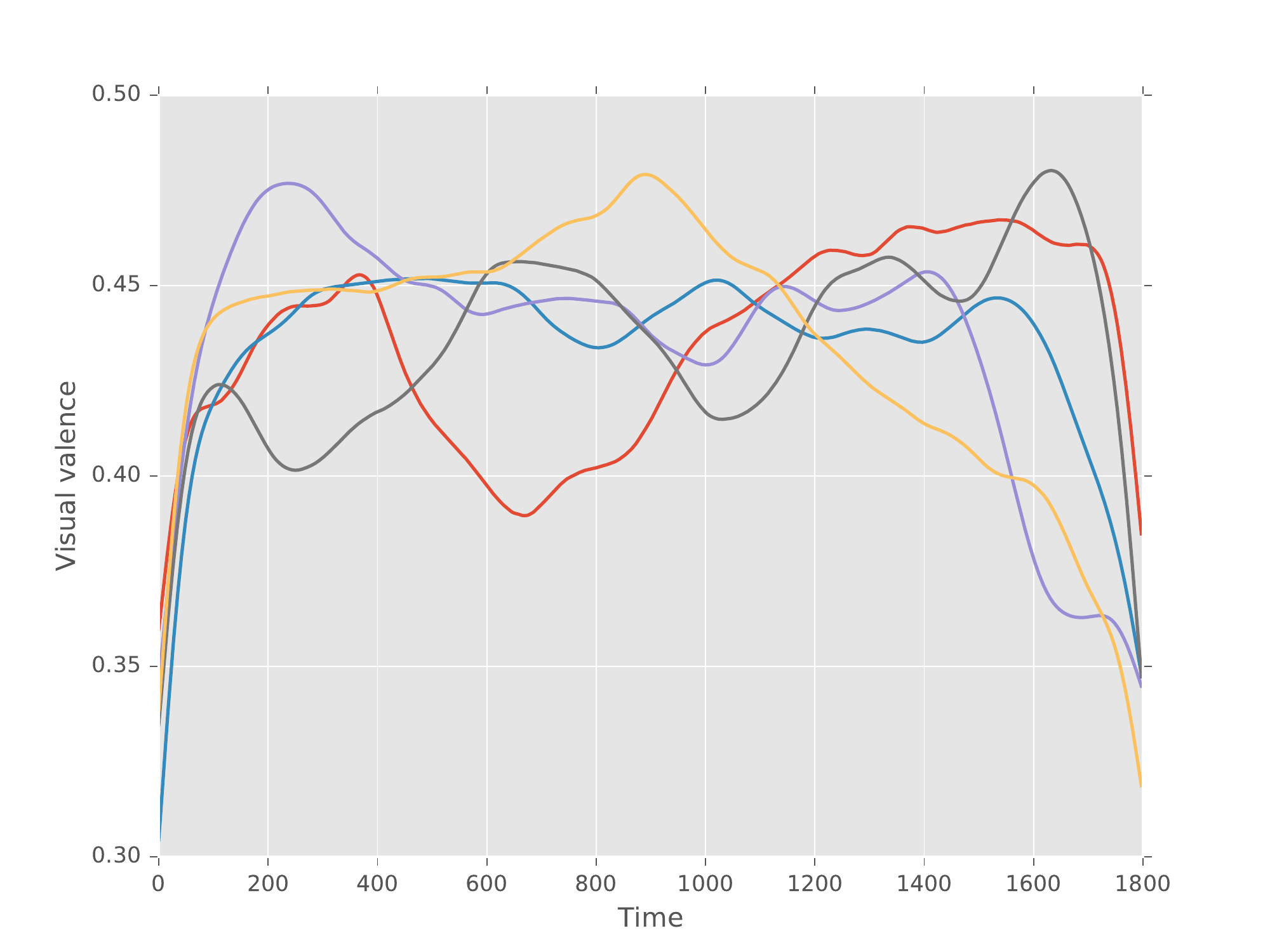}
	\caption{Clustering on Shorts Corpora with $k$ = 5, $w$ = 0.1}
	\label{kmedoids_shorts_k5}
\end{figure}

\vspace{-1em}
Figure \ref{kmedoids_shorts_k5} shows one example clustering, representing five typical emotional arcs. Note that the steep inclines and declines at the start and end are artifacts of opening scenes and credits. Compared to the Films Corpora, typical arcs in the Shorts Copora tend to be less complex, but also more extreme (e.g. the yellow arc that ends on a steady decline).

\vspace{-0.5em}
\subsection{Engagement analysis}\label{sec_engagement}
We can now return to the question of whether a video's emotional arcs affects the degree of viewer engagement.We performed a small experiment on our Shorts Corpora by using a) metadata features, and b) categorical cluster assignment features (which family of visual arcs does this movie belong to) as inputs to a regression model that predicts the number of comments a video received on Vimeo.

\begin{figure}[h!]
	\centering
	\includegraphics[width=0.9\linewidth]{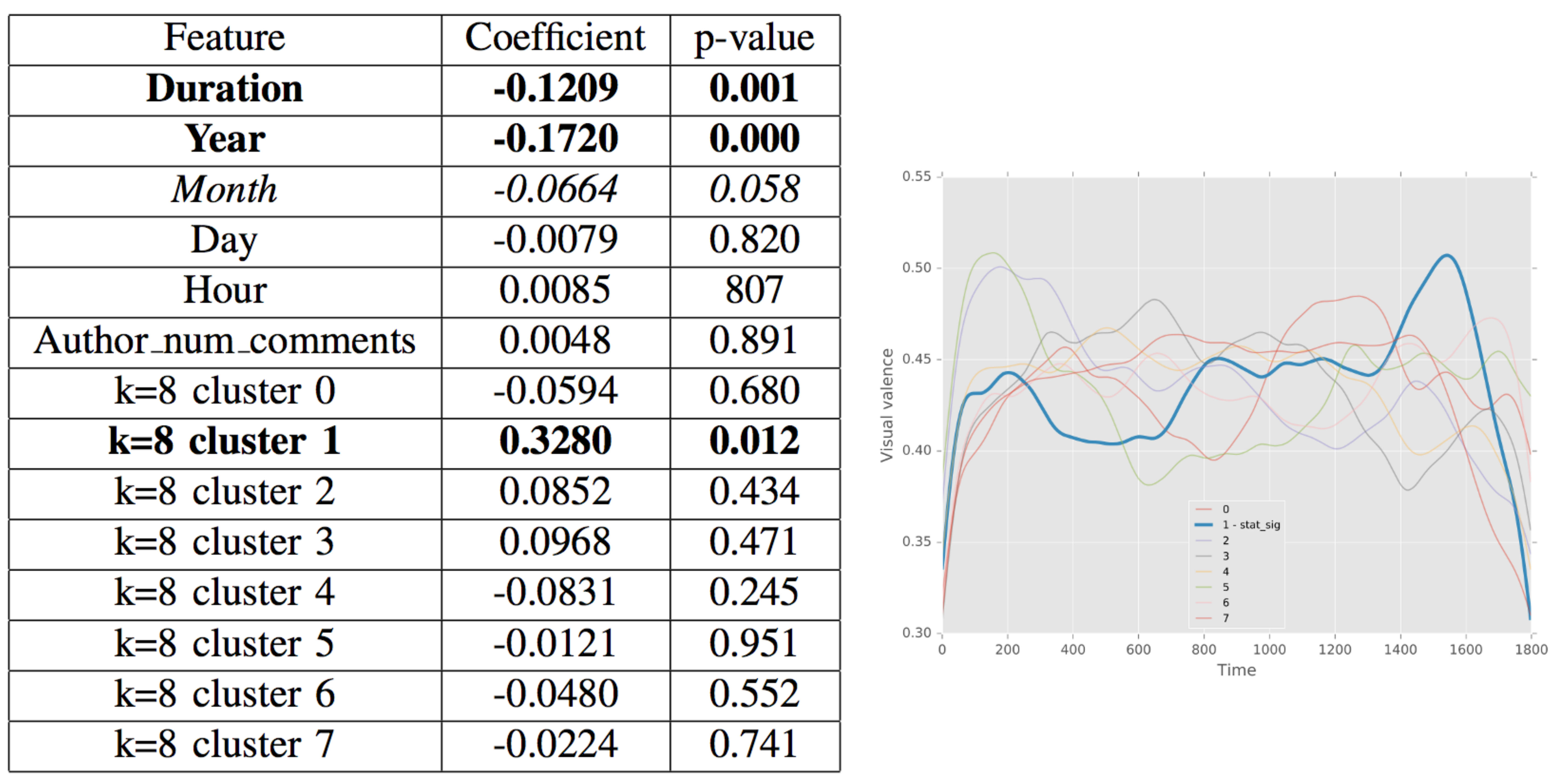}
	\caption{TABLE IX. Engagement analysis for $k=8$ clustering. P-values less than 0.05 are bolded; less than 0.1 are italicized. Statistically significant predictive arc is shown.}
	\label{consumption_k8}
\end{figure}

Nine models are created -- one for each value of $k$ in [2,10]. Not surprisingly, the duration and year are often stat-sig predictors. However, three arcs are also stat-sig, each positively correlated with the number of comments. The first stat-sig arc, the yellow arc shown in Figure \ref{kmedoids_shorts_k5}, fits the ``Icarus'' shape (rise-fall). The second ($k$ = 8) and third ($k$ = 10) arcs are characterized by a large peak near the end, with the former having some incline before the peak and the latter flat before the peak. In other words, they end with a bang. The results for $k$ = 8 are shown in Figure \ref{consumption_k8} and Table IX.
\vspace{-0.5em}

\vspace{-0.9em}
\section{Concluding Remarks}

We first developed methods for constructing and finding families of arcs. The crowdsourced annotation data prompts a number of possibilities for future work, such as dialogue-based arcs and plot-based sentiment modeling.

We were also able to show the predictive power of emotional arcs on a small subset of online Vimeo shorts. While intriguing, this was performed a) using only the visual arcs, and b) against a relatively simple metric. More data for the combined audio-visual model should generate more accurate arcs. It would also be interesting to see how, if at all, emotional features affect how videos propagate through social media sites like Twitter and Reddit.

\end{document}